\documentclass[letterpaper, 9 pt, conference]{ieeeconf}  

\IEEEoverridecommandlockouts

\usepackage{cite}
\usepackage{amsmath,amssymb,amsfonts}
\usepackage{algorithmic}
\usepackage{graphicx}
\usepackage{textcomp}
\usepackage{xcolor}
\usepackage{gensymb}
\usepackage{caption}
\usepackage{subcaption}
\usepackage{hyperref}
\usepackage{algorithm}

\usepackage{array}

\usepackage{listings}

\usepackage{booktabs}
\usepackage{multirow}
\usepackage{siunitx}

\hypersetup{
colorlinks=true,
urlcolor=blue}

\def\BibTeX{{\rm B\kern-.05em{\sc i\kern-.025em b}\kern-.08em
    T\kern-.1667em\lower.7ex\hbox{E}\kern-.125emX}}

\title{Interleaved POMDP Planning for Multi-Object Search in Unknown Multi-Room Household Environments}

\author{Ruochu Yang$^{1}$, Ziyi Xia$^{2}$, Huibo Zhang$^{2}$, Yatong Han$^{3}$, Yiming Zhao$^{3}$, \\ 
Yingke Li$^{4}$, Fumin Zhang$^{2}$, Yorai Wardi$^{1}$, Mengxue Hou$^{5}$
\thanks{$^{1}$Georgia Institute of Technology, USA. $^{2}$The Hong Kong University of Science and Technology, China. $^{3}$Ising AI, China. $^{4}$MIT, USA. $^{5}$Notre Dame University, USA.}
}

\begin{document}

\maketitle
\thispagestyle{empty}
\pagestyle{empty}

\begin{abstract}

Multi-object search in unknown household environments requires planning under extensive uncertainty — from unknown object locations to cluttered spaces with unobserved obstacles. POMDPs offer a principled framework for such problems but remain intractable in large domains. We propose Inter-POMDP, a novel interleaved POMDP planning algorithm that decomposes this challenge into two interacting levels: a high-level POUCT planner reasons over object distributions using LLM-informed histogram beliefs, while a low-level motion planner models navigation uncertainty with obstacle-aware particle beliefs as domain knowledge to guide high-level POUCT. This interleaved design balances planning quality and efficiency despite the large search space across unknown multi-room environments. Both simulation and real-world experiments show that our Inter-POMDP algorithm reduces collision counts by up to 63\%, navigation steps by up to 35\%, and detection counts by up to 32\% compared with baseline methods. Full videos are \url{https://sites.google.com/view/inter-pomdp}

\end{abstract}

\section{Introduction}
\label{intro}


Household robots are a long-standing topic in robotics, with object search being a classical problem. While many works address object search in deterministic settings, real-world deployments demand reasoning under partial observability: environments are large-scale, objects' locations vary over time, and obstacles are unknown. Planning under such uncertainty is critical for developing fully autonomous household robots. Two key challenges arise: 1) the large state-action space and long planning horizon demand efficiency, as the search tree is deep with many branching factors; 2) the extensive activity range in cluttered, uncertain environments makes obstacle encounters highly likely, demanding planning quality and robustness. Partially observable Markov decision processes (POMDPs) provide a principled framework for sequential decision-making under uncertainty \cite{kaelbling1998planning}. However, POMDPs are computationally intractable in large domains \cite{madani1999undecidability}, as belief space dimensionality grows with the number of possible states—exponentially so with the number of objects in search tasks \cite{pineau2006anytime}. When the environment is only partially observed, the full state-action space is unknown and vast, making long-horizon planning especially difficult.

We focus on cluttered multi-room household environments such as research laboratories or shared office workspaces, characterized by large open floors with similar furniture (e.g., tables, shelves) and miscellaneous objects (e.g., equipment, monitors, personal items). In such settings, digging more informative object-level and object-furniture relations is significant for efficient search \cite{chang2021comprehensive}. Target objects may rest on the ground, serving simultaneously as obstacles and search targets, necessitating the integration of high-level semantic information into low-level navigation costs—beyond approaches that only consider distance-based reachability \cite{zheng2022towards}. Moreover, large household environments emphasize an inevitable challenge: cluttered navigational space. Arbitrary arrangements of tables, chairs, and objects make it insufficient to simply partition rooms and expect obstacle-free navigation between them. We aim to systematically incorporate grounded navigation information into the belief update process to make planning under uncertainty more robust. To achieve this, we treat uncertainty as physical rather than purely semantic. Many existing works \cite{zheng2023asystem, wandzel2019multi, ge2024commonsense}  rely on high-level task planners to re-select exploratory directions upon failure, seldom accounting for physical-world uncertainties. Even works that acknowledge such uncertainties often resort to simple replanning \cite{zheng2021multi, giuliari2021pomp} or LLM-based action re-selection strategies \cite{remy2023semantically, fu2024scene}.

Several works have applied POMDP planning to object search and related tasks \cite{holzherr2021efficient}. Online POMDP solvers such as POMCP \cite{silver2010monte} and DESPOT \cite{somani2013despot} use sparse belief sampling with particles and Monte Carlo Tree Search (MCTS) to grow belief trees within limited planning time. POMCP has scaled to large POMDPs by sampling states from current beliefs and histories via a black-box simulator. However, its UCT-based exploration can be overly greedy with poor worst-case performance, and its black-box policies lack transparency, reducing safety and trustworthiness. Multi-scale approaches \cite{holzherr2021efficient} have shown that hierarchical methods maintain comparable solution quality on modest problems while demonstrating superior performance on larger ones. Our work is motivated by the interleaved planning workflow of \cite{hou2023interleaved}, which addresses MDP settings with known dynamics in pre-partitioned environments. We make a significant extension to POMDP planning, where transition models are unknown due to unobserved obstacles, and incorporate abstract semantic reasoning for efficient object search in multi-room environments, which is absent in prior MDP-based formulations. Our contributions are summarized as follows:
\begin{itemize}

\item We propose a novel interleaved POMDP planning framework for multi-object search in unknown multi-room environments. Our two-level planners handle semantic object distributions and geometric navigation, respectively, interacting iteratively so that low-level domain knowledge guides high-level action selection, balancing planning quality and efficiency against uncertainties from numerous unknown objects and obstacles.

\item Comprehensive evaluations demonstrate that our framework converges to near-optimal plans under realistic uncertainties such as furniture occlusions and unknown obstacles, leveraging search heuristics from past experience and natural language priors at both planning levels for safe and efficient operation.

\end{itemize}

\section{Related Works}
\label{related works}


\paragraph{Object Search in Unknown Environments}

In this general domain, there are a number of works solving this problem from different perspectives. 
\textit{Learning-based methods} train robots to explore and find objects through repeated environment interactions \cite{wani2020multion, ravichandran2022hierarchical}. However, they suffer from inefficient exploration—optimizing solely from sensory inputs without directly modeling semantic and geometric scene information \cite{chen2022learning} - and poor generalizability due to extensive training data requirements that hinder sim-to-real transfer \cite{sadek2023multi}.
\textit{Large foundation model (LFM) methods} leverage natural language and visual inputs for navigation in unknown environments. \cite{remy2023semantically} uses LLM semantic priors to guide online POMDP planning in scene graphs but ignores geometric constraints and past planning experience. \cite{fu2024scene} enhances interactive planning via 3D-visual-language models. \cite{ge2024commonsense} combines LLMs with scene graphs for target localization but assumes off-the-shelf navigation and relies heavily on room-level hierarchy, limiting adaptability to multi-room environments. Generally, LFM methods depend on coarse embeddings capturing only scene-level information, missing fine-grained object correlations and physical constraints \cite{gadre2023cows}.
\textit{Probabilistic planning methods} systematically account for uncertainty using frameworks like POMDPs, typically assuming no prior environment knowledge \cite{zheng2023asystem, wandzel2019multi, silver2010monte}. While principled, POMDP methods face computational challenges in scaling to large environments due to maintaining belief states for multiple objects over extended horizons \cite{zheng2021multi, giuliari2021pomp}. In this work, we focus on the POMDP domain, deliberately addressing the challenges and limitations of existing methods for object search.

\paragraph{POMDP Planners}

We focus on POMDP planning for object search. \cite{giuliari2021pomp} introduces belief reinvigoration for dynamically growing state spaces in online map generation. \cite{zheng2023asystem} builds a generalized 3D multi-object search system but uses naive POMCP as a plug-in without adapting it to the problem structure. \cite{chen2024pomdp} proposes GPOMCP for finding occluded objects but ignores navigation uncertainty and semantic guidance. \cite{wandzel2019multi} formulates multi-object search as an OO-POMDP with OO-POMCP, but only utilizes semantics at task initialization; we continuously exploit semantic information throughout the mission. \cite{zheng2021multi} extends to 3D with multi-resolution planning but assumes object independence, ignores geometric constraints in observations, and lacks rollout heuristics. \cite{amiri2022reasoning} incrementally constructs scene graphs but focuses on representation rather than planning, relying on off-the-shelf POMDP solvers. \cite{jamgochian2024constrained} scales constrained-POMDP planning via hierarchical MCTS but does not leverage semantic information for household object search. Most related to our work, \cite{zheng2022towards} builds a hierarchical framework with two POUCT planners. However, it follows a purely sequential decision paradigm without feeding low-level navigation costs back to the high-level planner, its factored distribution reverts to an intractable joint distribution as targets increase, and its learned correlation model lacks generalizability. To the best of our knowledge, we propose the first interleaved POMDP planning algorithm for multi-object search in unknown multi-room environments, designing a principled interaction between two planners to jointly capture semantic and geometric information under uncertainty.

\section{Problem Formulation}
\label{problem formulation}


We consider a large-scale household environment with $K$ known furniture items $Fur = \{fur_1, \ldots, fur_K\}$. A mobile robot equipped with an onboard camera is tasked with finding $N$ unknown target objects $O_{tar} = \{tar_1, \ldots, tar_N\}$ . The environment additionally contains $M$ unknown landmark objects $O_{obj} = \{obj_1, \ldots, obj_M\}$ that may provide contextual cues, and $J$ unknown obstacles $O_{obs} = \{obs_1, \ldots, obs_J\}$ on the floor that may impede navigation. The robot has access to a 2D occupancy grid map $M^o$ encoding walls and furniture, but has no prior knowledge of target objects, landmark objects, or obstacles. Cells outside the robot's field of view are initialized as unknown and updated incrementally during exploration. Importantly, we explicitly model object-object and object-furniture relations, departing from the independence assumption in OO-POMDP \cite{wandzel2019multi}.

At each time step $t$, the joint state is: $s_t = \{s_t^r, S_{O_{tar}}, S_{O_{obj}}, S_{O_{obs}}\} \in S$, where the robot state $s_t^r = (x_t^r, y_t^r, \theta_t^r)$ specifies its position and orientation on $M^o$ and is fully known. Each target object state $s_t^{tar_i} = (x_t^{tar_i}, y_t^{tar_i}, A^{tar_i})$ and landmark object state $s_t^{obj_j} = (x_t^{obj_j}, y_t^{obj_j}, A^{obj_j})$ include grid coordinates and semantic attributes. Each obstacle state $s_t^{obs_k} = (x_t^{obs_k}, y_t^{obs_k})$ specifies only its position. All non-robot states are unknown. Since $s_t$ is partially observable, the robot maintains a belief $b_t(s_t) = \Pr(s_t \mid h_t)$ over the joint state, updated via Bayes' Rule after each action and observation. The high dimensionality of this belief - growing with the number of unknown objects and obstacles - motivates a hierarchical decomposition.

\paragraph{Hierarchical POMDP Decomposition}

The joint POMDP is intractable for any single planner due to the large number of unknown entities. We decompose it into two levels: a high-level planner for semantic reasoning over object search order, and a low-level planner for geometric reasoning over navigation.

\textbf{High-Level Action Space $\mathcal{A}_h$:} Four abstract actions: \texttt{Navigate} moves the robot between regions; \texttt{Search} moves the robot in a local region; \texttt{Sense} performs object detection within the robot's field of view (FOV); and \texttt{Declare} terminates the mission, succeeding if all targets are found.

\textbf{Low-Level Action Space $\mathcal{A}_l$:} To execute \texttt{Navigate} or \texttt{Search}, the robot takes primitive actions \texttt{{move\_ahead, turn\_left, turn\_right}}, and to execute \texttt{Sense} with camera taking pictures.

\textbf{Transition Model $T$:} \texttt{Navigate} action moves the robot between grid cells on $M^o$. Crucially, unknown obstacles $O_{obs}$ may lie along the path, introducing trajectory-level uncertainty. We account for this by reasoning over \texttt{Navigate} at a higher-resolution trajectory space - going beyond prior works \cite{chen2024pomdp, remy2023semantically} that consider only high-level search while ignoring grounded navigation constraints.

\textbf{Observation Space $Z$ and Model $H$:} Upon executing \texttt{Sense}, the robot receives observation $z_t = (z_t^{tar_i}, z_t^{obj_j}, z_t^{obs_k})$, where each component is either the detected state within the FOV or $null$ if undetected. The observation model $H(z_t, s_t^r) = \Pr(z_t \mid s_t^r)$ is defined by a fan-shaped camera of depth $D$ with perfect accuracy, as object detection is outside the scope of this work.

\textbf{Reward Model $R$:} A successful \texttt{Declare} (all targets detected within FOV and within a distance threshold) yields $R = +100$; failure yields $R = -100$. Each \texttt{Navigate} action incurs $R = -1$, and each collision with an obstacle incurs $R = -10$.

\paragraph{Object Relations and Enhanced Maps}

To exploit the rich structure of household environments, we introduce two additional elements beyond the standard POMDP formulation.

\textbf{Object-Furniture Relations:} We define object-object relations $C^{oo}(s^{tar_i}, s^{obj_j}) = \Pr(s^{tar_i} \mid s^{obj_j})$ (e.g., \textit{coffee\_cup next\_to milk}) and object-furniture relations $C^{of}(s^{tar_i}, s^{fur_p}) = \Pr(s^{tar_i} \mid s^{fur_p})$ (e.g., \textit{coffee\_cup on kitchen\_table}). These relations encode semantic priors that guide the high-level planner toward promising search locations.

\textbf{Enhanced Maps:} Beyond the occupancy grid $M^o$, we maintain two complementary representations: (1) a \emph{semantic map} $M^s$ that abstracts grid cells to furnitures $fur_p$, incrementally built online using relations $C^{oo}$ and $C^{of}$ to inform high-level search decisions; and (2) a \emph{topological map} $M^p = \langle \mathcal{N}, \mathcal{E}, \mathcal{W} \rangle$, a sparse graph over reachable positions where edge weights $\mathcal{W}$ encode navigation costs arising from unknown obstacles $O_{obs}$, incrementally updated as the robot explores to inform low-level planning.

\paragraph{Problem Statement}

Given the above formulations, our goal is to find an optimal policy $\pi^*$ that guides the robot to find all target objects $O_{tar}$ while minimizing cumulative cost. This problem is NP-hard: it requires jointly optimizing the high-level search order over objects and the low-level navigation under obstacle uncertainty, across a large environment with an expansive solution space. The central challenge is: \emph{how can we design a POMDP planning algorithm that achieves a principled balance between planning quality and efficiency under such complexity?}

\section{Methodology}
\label{methodology}

\begin{figure*}
        \centerline{\includegraphics[width=\textwidth]{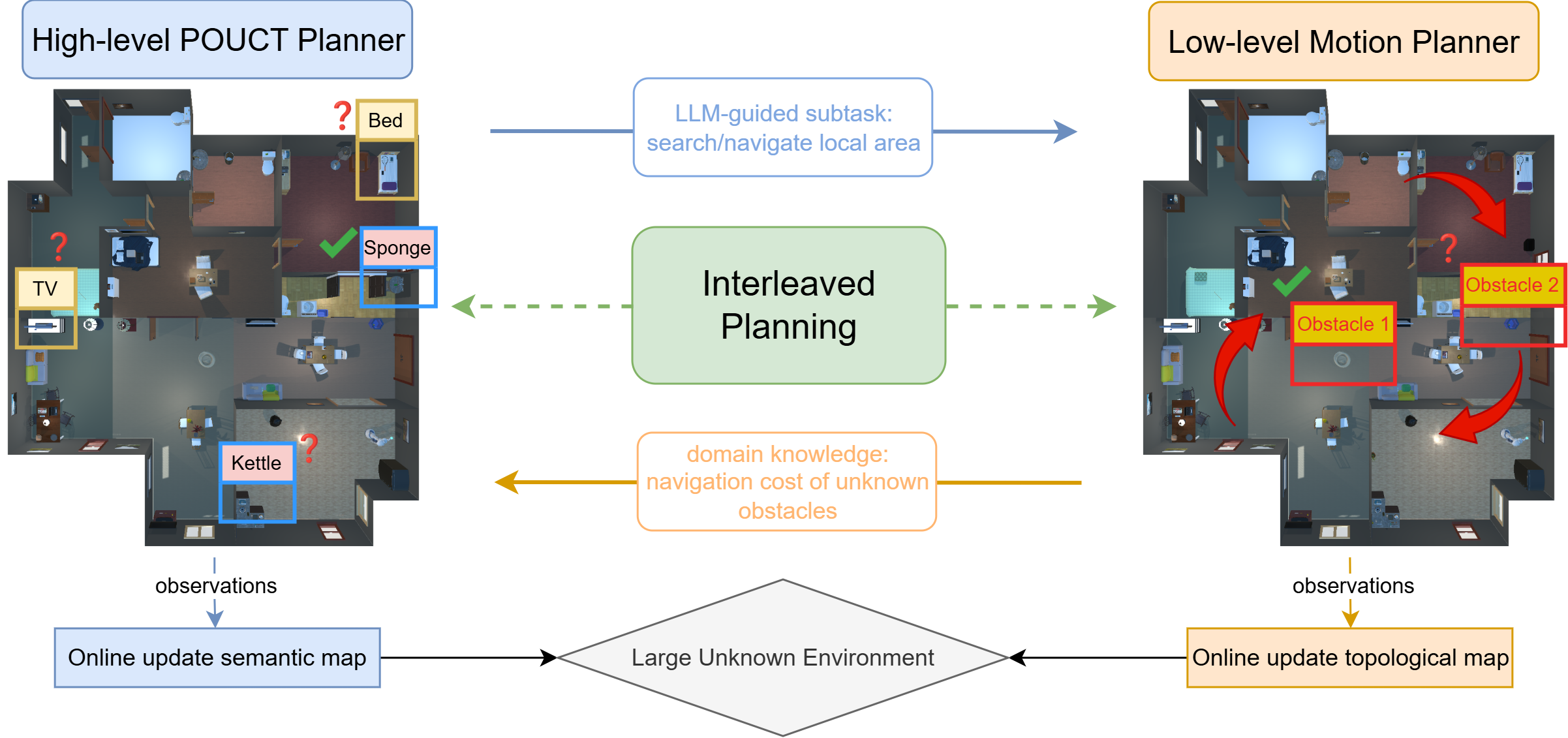}}
        \caption{Interleaved POMDP planning for large unknown environments. The high-level POUCT planner issues LLM-guided semantic subtasks to the low-level motion planner, which resolves local obstacles and feeds navigation costs back as domain knowledge. Both planners continuously update their semantic and topological maps from online observations.}
        \label{main method}
\end{figure*}

As shown above, our problem formulation introduces many challenges which traditional POMDP planners can't directly solve. To this end, we propose a novel interleaved POMDP planning algorithm to systematically imbue high-level semantic information and low-level geometric information with two specific designs:
\begin{itemize}

    \item Structural planning hierarchy to mitigate the curse of dimensionality where the high-level planner deals with global object localization and the low-level planner deals with motion primitives. 

    \item  Interleaved communication between the two planners where the high-level planner narrows the search space for the motion layer, while the low-level planner provides grounded navigation feedback to prune redundant search branches.
    
\end{itemize}

During a long process of searching multiple objects, our interleaved planning algorithm will incrementally model the large unknown environment, starting from an empty map with only furniture to a detailed map with all the objects/obstacles included. Our algorithm considers both quick target localization and robust navigation with unknown obstacles. In this way, we can enjoy a balance of efficiency and accuracy, which is the best we can expect to plan on multi-object search in a large unknown environment with obstacles. In this section, we first introduce our hierarchical architecture of two POMDP planners. Then we present our interleaved design of connecting the two planners in a principled way. Details are described in Algorithm \ref{main algo}.

\begin{algorithm}
\caption{Interleaved POMDP Planning}
\label{main algo}
\begin{algorithmic}[1]
\REQUIRE initial state $s_0$, target object $tar$, initial observation $z_0$, action set $\mathcal{A}$, occupancy grid map $M^o_t$, semantic map $M^s_t$, topological map $M^p_t = \langle \mathcal{N}, \mathcal{E}, \mathcal{W} \rangle$.
\ENSURE  high-level subtask $a_{high}$, low-level action $a_{low}$,  updated maps  $M^o_{t+1}$, $M^s_{t+1}, M^p_{t+1}$.

\WHILE{$tar$ not reached}
    \STATE \textbf{// High-Level POUCT Planning}
    \STATE Update semantic map $M^s_t$ using object-furniture spatial relations.
    \STATE $a_{high} \leftarrow$ Select semantic subtask via POUCT based on $M^s_t$ and belief.
    
    \STATE \textbf{// Low-Level Motion Planning}
    \STATE Sample obstacle states $s_t^{obs}$ into particle belief representations.
    \STATE Update topological nodes $\mathcal{N}$ and edges $\mathcal{E}$ based on $s_t^{obs}$ and $M^o_t$.
    \STATE Compute navigation costs $c$ via A* path planning on $M^o_t$
    \STATE Update edge weights $\mathcal{W}$ in $M^p_t$ using computed costs $c$.
    
    \STATE \textbf{// Interleaved Planning}
    \STATE Propagate low-level navigation costs $c$ back to High-level POUCT as domain knowledge.
    \STATE $a_{low} \leftarrow$ Execute motion primitive to satisfy $a_{high}$ given updated costs.
    
    \STATE \textbf{// Bi-level Belief Update}
    \STATE Update $b_{t+1} \leftarrow b_t$ based on robot state $s^r_t$ and environmental observation $z_t$.
    \STATE High-level POUCT updates $M^s_{t+1} \leftarrow M^s_t$ and Low-level planner updates  $M^o_{t+1} \leftarrow M^o_t$ and $M^p_{t+1} \leftarrow M^p_t$,
\ENDWHILE
\end{algorithmic}
\end{algorithm}

\subsection{High-Level POUCT for Generalized Object Localization}

The high-level planner operates on the semantic map $M^s$ to identify promising search locations for target objects. Its core task is to select which furniture or region to visit next, based on inferred object-object relations $C^{oo}$ and object-furniture relations $C^{of}$.

\begin{figure}
        \centerline{\includegraphics[width=0.5\textwidth]{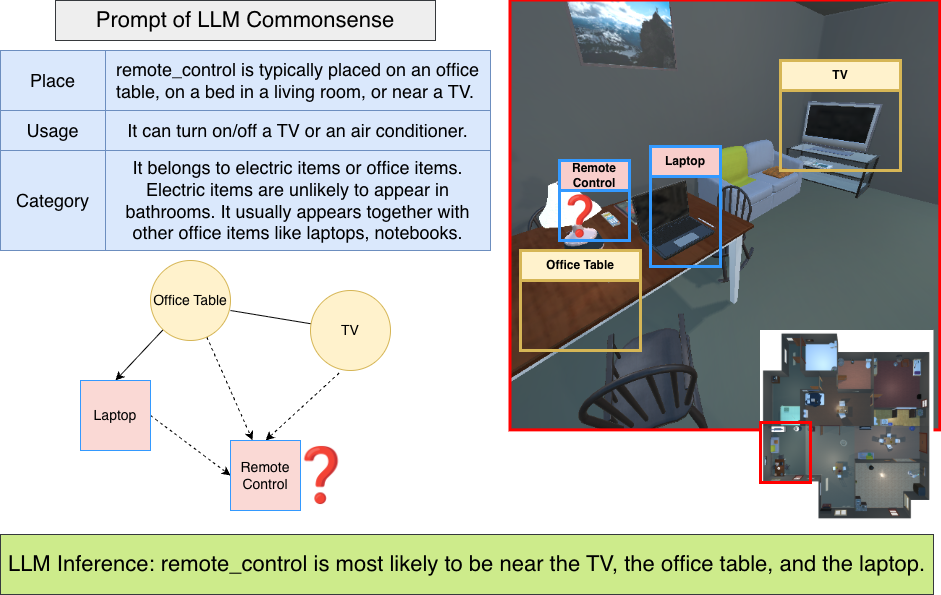}}
        \caption{LLM prompt design of generalized object localization for the high-level POUCT planner based on the explored objects and the pre-known furniture.}
        \label{llm prompt}
\end{figure}

\textbf{LLM Relation Inference:}
Modeling object relations in advance is infeasible in large environments with many unknown objects, and training relation models requires extensive data to generalize across diverse scenes. We instead leverage LLMs, which possess commonsense knowledge that naturally serves as semantic priors for search. To fully exploit LLM reasoning, we encode semantic attributes $A^{obj} = \texttt{{location, category, usage}}$ for each object. For example, \texttt{location}: a cup is typically near food and drinks on a table; \texttt{usage}: a cup is used for holding drinks during leisure activities. The relation models are computed via structured prompts:
\begin{equation}
C^{oo}, C^{of} \leftarrow \text{LLM}(Fur, O_{obj}, O_{tar}).
\end{equation}
To capture environment-specific preferences (e.g., a user who always places their coffee cup on the study table), we additionally maintain a memory map $M_m$ that allows the LLM to accumulate domain-specific knowledge online. LLM prompt design is shown in Figure \ref{llm prompt}.

\textbf{Histogram Belief Representation:}
We represent the high-level belief as a histogram over discretized furniture-level locations, enabling exact Bayesian inference. The planner maintains two complementary beliefs: a \emph{primary belief} $b$, updated from POMDP observations, and a \emph{refined belief} $b'$, obtained by conditioning $b$ on the semantic relations $C^{oo}$ and $C^{of}$. As the robot explores and discovers landmark objects, these relations help narrow the search region - if a landmark frequently co-occurs with a target, finding it significantly constrains the target's likely location. Given the current semantic map $M^s_t$, beliefs, and relation models, the high-level POUCT selects a subtask:
\begin{equation}
\textit{subtask} \leftarrow \text{POUCT}(M^s_t, b', C^{oo}, C^{of}).
\end{equation}

\subsection{Low-Level Motion Planner for Uncertainty-aware Navigation}

While the high-level planner determines \emph{where} to search, the low-level planner determines \emph{how} to get there safely. It reasons over fine-grained geometric uncertainties arising from unknown obstacles using particle-based beliefs.

\textbf{Particle Belief Representation:}
Given a high-level subtask (\texttt{Navigate} or \texttt{Sense}), the low-level POMCP planner maintains a particle belief over obstacle states ${s^{obs_k}}$, updated via the transition model $s_{t+1} = T^l(s_t, a_t)$ and observation model $p(z_t \mid s_{t+1})$. Each particle represents a hypothesis about the local obstacle configuration, enabling the planner to reason about collision risks along candidate trajectories.

\textbf{Topological Map Construction:}
The planner maintains a topological map $M^p_t = \langle \mathcal{N}, \mathcal{E}, \mathcal{W} \rangle$ for grounded navigation. Nodes $\mathcal{N}$ are reachable positions sampled from current beliefs over targets $b_t(s_t^{tar_i})$ and obstacles $b_t(s_t^{obs_k})$. Edges $\mathcal{E}$ ensure graph connectivity, and weights $\mathcal{W}$ encode navigation costs influenced by discovered and hypothesized obstacles. As the robot explores, new observations update the particle belief and consequently refine $M^p_t$, yielding increasingly accurate navigation cost estimates.

\textbf{Navigation Cost as Domain Knowledge:}
Since the local geometry is initially unknown and the trajectory space is vast, the low-level planner cannot guarantee optimal paths from the outset. However, as it accumulates planning experience—successful traversals, observed obstacles, and collision events—it produces increasingly reliable navigation cost estimates. These costs constitute valuable domain knowledge that, if communicated to the high-level planner, can significantly improve global planning quality. This observation motivates the interleaved design described next.

\subsection{Interleaved POMDP Planning with Domain Knowledge}

Given the above hierarchy, a key question remains: \emph{how should the two planners interact so that the overall policy optimally hedges against both semantic uncertainty in object locations and geometric uncertainty in navigation?} Most hierarchical POMDP approaches operate sequentially \cite{giuliari2021pomp} or without interaction
\cite{holzherr2021efficient, zheng2022towards} - the high-level planner issues subtasks without knowledge of their execution cost, and the low-level planner executes them without influencing future high-level decisions. This separation can lead to suboptimal plans: the high-level planner may repeatedly select subtasks that are semantically promising but geometrically costly (e.g., navigating through heavily obstructed areas), without any mechanism to learn from these experiences. Our key innovation is an interleaved mechanism where low-level navigation costs are propagated upward as domain knowledge to bias high-level action selection, enabling the two planners to iteratively co-refine the global plan.

\textbf{Domain Knowledge Augmented UCT:}
POUCT planners treat action selection at each tree node as a multi-armed bandit problem, using UCT \cite{auer2002finite} to balance exploration and exploitation:
\begin{equation}
Q^\oplus(s, a) = Q(s, a) + c \sqrt{\frac{\log N(s)}{N(s,a)}},
\end{equation}
where $Q(s,a)$ is the estimated action value, $N(s,a)$ the visit count, and $c$ the exploration constant. In standard MCTS, $Q(s,a)$ is initialized to zero, providing no guidance early in the search. We instead initialize it using domain knowledge from the low-level planner:
\begin{equation}
Q(s, a) \leftarrow Q_{\text{init}}(s, a),
\end{equation}
where $Q_{\text{init}}(s,a)$ encodes accumulated navigation costs from low-level geometric reasoning. This initialization biases the tree search toward states that are both semantically promising and geometrically feasible, without altering the asymptotic convergence guarantees of MCTS. Similarly, the rollout policy incorporates this domain knowledge to select informed actions rather than sampling uniformly at random, further accelerating convergence to high-quality plans.

\section{Experiments}
\label{experiments}

We evaluate our Inter-POMDP and compare with two baselines CSG-TL \cite{ge2024commonsense} and COSPOMDP \cite{zheng2022towards} on multi-object search in unknown multi-room environments with narrow paths and unknown obstacles.

\subsection{Experimental Setup}

\textbf{Simulation:} We use ProcTHOR \cite{procthor} of 8-12 rooms with numerous objects/furnitures and a mobile robot with an onboard camera.

\textbf{Real-World:} We deploy on a RealMan RMC-DA 7DOF dual-arm mobile robot with STM32 chassis and Nvidia Jetson AGX Orin compute platform in an $8.71\text{m} \times 7.22\text{m}$ room and YOLO11 \cite{khanam2024yolov11} for object detection.

\textbf{Baselines:} We compare against methods CSG-TL \cite{ge2024commonsense} and COSPOMDP \cite{zheng2022towards}. We aim to demonstrate that our interleaved design can enable bi-directional information flow between planning levels for globally informed plans.

\textbf{Metrics:} We evaluate three grounded metrics: \emph{collision counts}, \emph{navigation steps}, and \emph{detection counts}. Results are reported as mean $\pm$ standard deviation across 5 trials per mission.

\textbf{Configurations:} We use GPT-4o \cite{hurst2024gpt} for LLM relation inference. We use GATv2Conv \cite{brody2021attentive} backbone for relation model training with hyperparameters as 300 epochs, 16 batch size, $1 \times 10^{-4}$ learning rate, Adam optimizer, and binary cross-entropy loss function and with training dataset as 1500 ProcTHOR object/furniture relations.

\subsection{Simulation Experiments}

\begin{figure*}
        \centerline{\includegraphics[width=\textwidth]{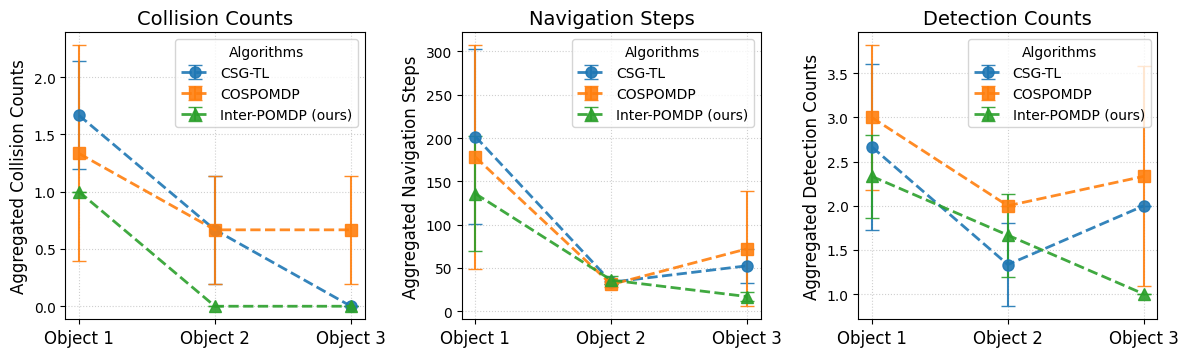}}
        \caption{Aggregated performance comparison  across unknown multi-room ProcTHOR scenes train\_1, train\_8, and train\_13  between CSG-TL \cite{ge2024commonsense}, COSPOMDP \cite{zheng2022towards}, and Inter-POMDP (ours) in terms of collision counts, navigation steps, and detection counts.}
        \label{fig:aggregated}
\end{figure*}

\begin{figure*}
    \centering
    
    \begin{subfigure}[b]{\textwidth}
        \includegraphics[width=\textwidth, height=7cm]{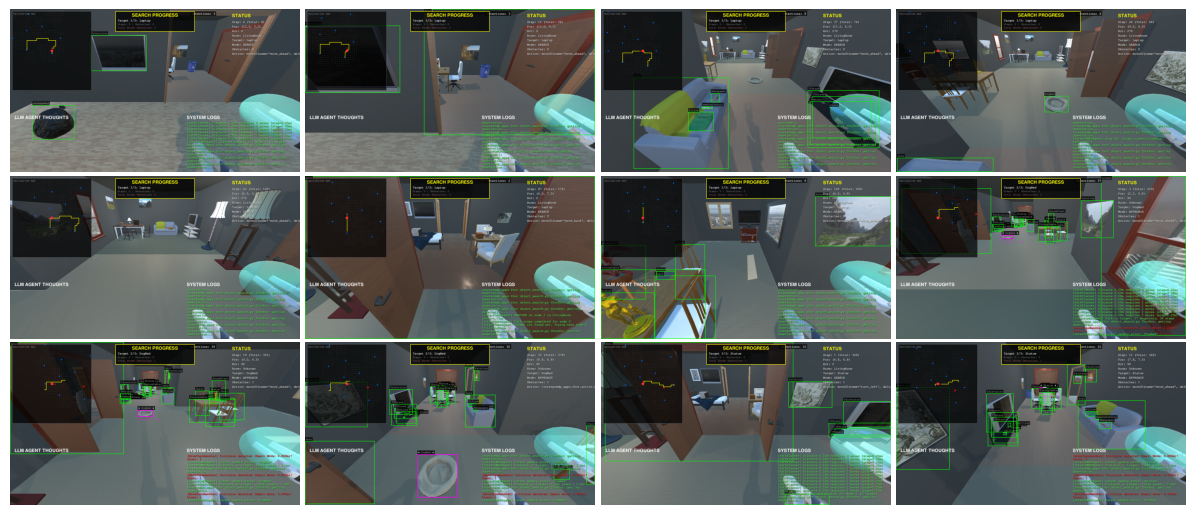}
        \caption{CSG-TL \cite{ge2024commonsense} combines LLMs with scene graphs for target localization but assumes off-the-shelf navigation and ground-truth room-level hierarchy. As a result, it incurs more obstacle collisions and requires more navigation steps when facing narrow aisles and cluttered spaces.}
        \label{fig:experiments/simulation/train_1_csg_only}
    \end{subfigure}
  
     \vspace{-0.4em}
    
    \begin{subfigure}[b]{\textwidth}
        \includegraphics[width=\textwidth, height=7cm]{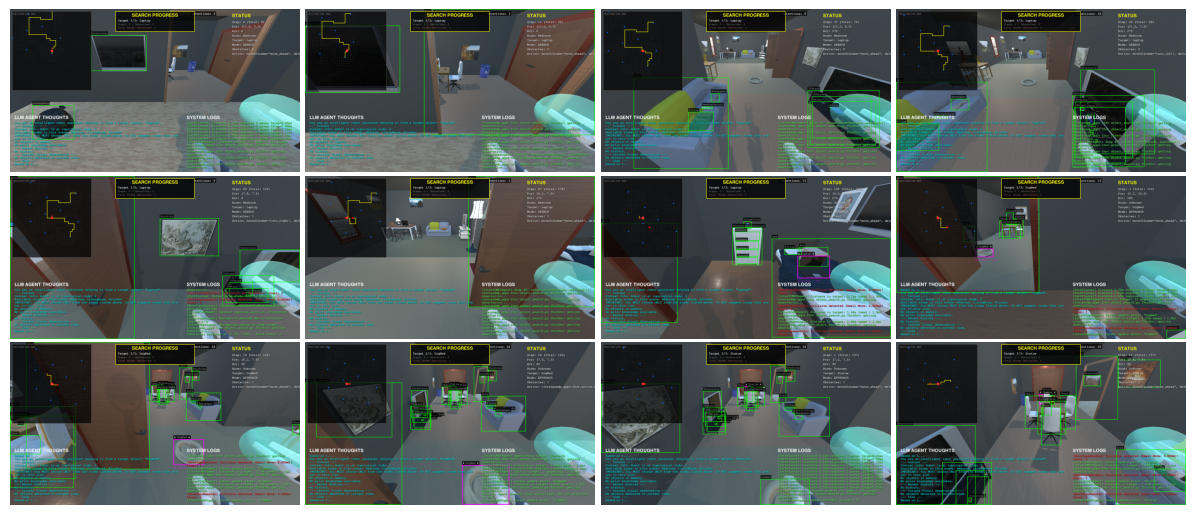}
        \caption{COSPOMDP \cite{zheng2022towards} is limited by a naive correlation model that lacks generalizability for multi-room, multi-object settings, and histogram beliefs without low-level navigation feedback that lack fidelity for fine-grained planning.}
        \label{fig:experiments/simulation/train_1_llm_only}
    \end{subfigure}
    
    \vspace{-0.4em}

    \begin{subfigure}[b]{\textwidth}
        \includegraphics[width=\textwidth, height=7cm]{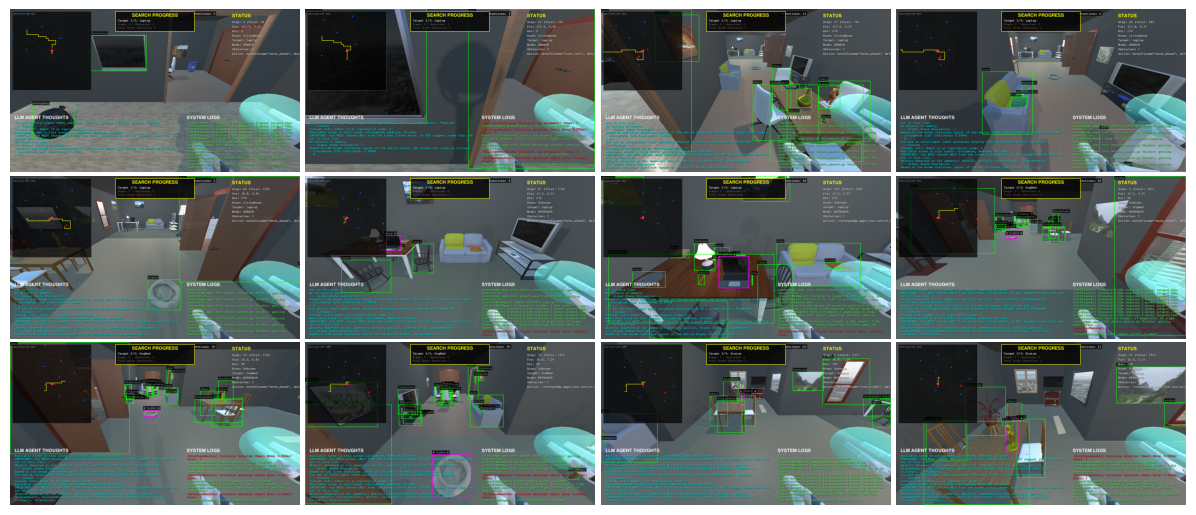}
        \caption{Our Inter-POMDP algorithm finds all three target objects more quickly while colliding fewer obstacles and requiring fewer camera activations.}
        \label{fig:experiments/simulation/train_1_interpomdp}
    \end{subfigure}

     \vspace{-0.4em}

    \caption{Qualitative comparison in ProcTHOR train\_1 scene with aligned timestamps. Full videos are \url{https://sites.google.com/view/inter-pomdp}}
    \label{fig:simulation}
    
\end{figure*}

We evaluate all methods across 15 trials in three large ProcTHOR scenes train\_1, train\_8, and train\_13, where each trial requires the robot to find 3 target objects across 8-12 rooms. Quantitative results are shown in Table \ref{table:all_metrics}, aggregated performance is shown in Figure \ref{fig:aggregated}, and qualitative comparisons with aligned timestamps are shown in Figure \ref{fig:simulation}. An obvious trend is that Inter-POMDP's performance achieves progressively the strongest results across all scenes: zero collisions, fewest navigation steps and camera detection. This validates our interleaved design - propagating low-level domain knowledge back to the high-level for near-optimal planning by the end.

\begin{table*}
\caption{Comprehensive metric results of algorithm comparison between CSG-TL \cite{ge2024commonsense}, COSPOMDP \cite{zheng2022towards}, and Inter-POMDP (ours). Totally we conduct 15 trials in 3 large ProcTHOR scenes where each one requires the robot to search 3 objects in mutilple rooms. Full videos are \url{https://sites.google.com/view/inter-pomdp} }
\vspace{-2mm}
\begin{center}
\setlength{\tabcolsep}{5pt}
\renewcommand{\arraystretch}{1.5}
\begin{tabular}{| c | c || c | c | c | c | c | c | c | c | c|}

    \hline
    \multirow{3}{*}{\textbf{Metrics}} & \multirow{3}{*}{\textbf{Algorithms}}  & \multicolumn{9}{c|}{\textbf{15 Trials in 3 Large ProcTHOR Scenes}}  \\

    \cline{3-11}

     & &  \multicolumn{3}{c|}{\textbf{scene\_1}} & \multicolumn{3}{c|}{\textbf{scene\_8}} & \multicolumn{3}{c|}{\textbf{scene\_13}} \\

       \cline{3-11}

     & & \textbf{object\_1} & \textbf{object\_2} & \textbf{object\_3} & \textbf{object\_1} & \textbf{object\_2} & \textbf{object\_3} & \textbf{object\_1} & \textbf{object\_2}  & \textbf{object\_3} \\

    \hline

\multirow{3}{*}{Collision Counts} 
& CSG-TL \cite{ge2024commonsense}
& $2 \pm 0.1$ & $1 \pm 0.2$ & $0 \pm 0.1$ & $1 \pm 0.1$ & $0 \pm 0.1$ & $0 \pm 0.2$ & $2 \pm 0.2$ & $1 \pm 0.1$ & $0 \pm 0.1$ \\

& COSPOMDP \cite{zheng2022towards}
& $2 \pm 0.1$ & $1 \pm 0.1$ & $0 \pm 0.2$ & $0 \pm 0.1$ & $0 \pm 0.2$ & $1 \pm 0.1$ & $2 \pm 0.1$ & $1 \pm 0.1$ & $1 \pm 0.1$ \\

& \textbf{Inter-POMDP (ours)}
& $1 \pm 0.1$ & $0 \pm 0.1$ & $\mathbf{0 \pm 0.1}$ & $1 \pm 0.1$ & $0 \pm 0.1$ & $\mathbf{0 \pm 0.1}$ & $1 \pm 0.1$ & $0 \pm 0.1$ & $\mathbf{0 \pm 0.1}$ \\

   \cline{1-11}

\multirow{3}{*}{Navigation Steps} 
& CSG-TL \cite{ge2024commonsense} 
& $313 \pm 3$ & $34 \pm 2$ & $34 \pm 3$ & $223 \pm 4$ & $31 \pm 3$ & $43 \pm 3$ & $69 \pm 4$ & $35 \pm 1$ & $80 \pm 2$ \\

& COSPOMDP \cite{zheng2022towards}
& $108 \pm 3$ & $27 \pm 2$ & $26 \pm 2$ & $359 \pm 5$ & $31 \pm 3$ & $24 \pm 3$ & $67 \pm 5$ & $34 \pm 2$ & $166 \pm 5$ \\

& \textbf{Inter-POMDP (ours)}
& $101 \pm 3$ & $34 \pm 2$ & $\mathbf{12 \pm 1}$ & $229 \pm 3$ & $31 \pm 2$ & $\mathbf{24 \pm 2}$ & $77 \pm 3$ & $42 \pm 2$ & $\mathbf{14 \pm 1}$ \\

 \cline{1-11}

\multirow{3}{*}{Detection Counts} 
& CSG-TL \cite{ge2024commonsense} 
& $4 \pm 0.2$ & $2 \pm 0.2$ & $2 \pm 0.2$ & $2 \pm 0.2$ & $1 \pm 0.1$ & $2 \pm 0.1$ & $2 \pm 0.2$ & $1 \pm 0.1$ & $2 \pm 0.1$ \\

& COSPOMDP \cite{zheng2022towards}
& $3 \pm 0.3$ & $2 \pm 0.1$ & $2 \pm 0.2$ & $4 \pm 0.1$ & $2 \pm 0.1$ & $1 \pm 0.1$ & $2 \pm 0.1$ & $2 \pm 0.2$ & $4 \pm 0.3$ \\

& \textbf{Inter-POMDP (ours)}
& $3 \pm 0.2$ & $2 \pm 0.1$ & $\mathbf{1 \pm 0.1}$ & $2 \pm 0.2$ & $1 \pm 0.2$ & $\mathbf{1 \pm 0.2}$ & $2 \pm 0.1$ & $2 \pm 0.2$ & $\mathbf{1 \pm 0.1}$ \\

  \hline

\end{tabular}

\label{table:all_metrics}

\end{center}
\vspace{-2mm}
\end{table*}

\subsection{Analysis}

We perform detailed analysis between our Inter-POMDP and CSG-TL \cite{ge2024commonsense} and COSPOMDP \cite{zheng2022towards} based on quantitative metrics.

\textbf{Navigation Efficiency:}
For the first object, all methods require comparable steps as the environment is largely unexplored. The gap widens significantly for later objects: in train\_1, Inter-POMDP finds the third object in $12 \pm 1$ steps versus $34 \pm 3$ (CSG-TL) and $26 \pm 2$ (COSPOMDP); in train\_13, the contrast is more obvious at $14 \pm 1$ versus $80 \pm 2$ and $166 \pm 5$, a reduction of over 90\% and 80\%, respectively. As shown in Figure~\ref{fig:aggregated}, Inter-POMDP achieves the lowest aggregated navigation steps across all scenes. The baselines' inefficiency stems from distinct limitations: CSG-TL assumes off-the-shelf navigation and relies on room-level abstraction, missing fine-grained spatial nuances in geometrically complex scenes such as narrow aisles; COSPOMDP's naive correlation model lacks generalizability - occasionally degrading search efficiency as evidenced by its high step count in train\_13 - and its histogram beliefs at both levels without low-level feedback lack fidelity for fine-grained trajectory planning. In contrast, Inter-POMDP's interleaved mechanism effectively translates accumulated low-level navigation knowledge into more informed high-level subtask selection, yielding robust and efficient planning across diverse environments.

\textbf{Collision Avoidance via Particle Beliefs:}
Inter-POMDP consistently achieves fewer or equal collisions, reaching zero for the second and third objects across all scenes with a clear decreasing pattern ($1 \to 0 \to 0$), while both baselines exhibit higher counts without systematic reduction. This improvement stems from our use of particle-based beliefs with reinvigoration at the low level rather than use histogram representations at all levels, particle beliefs capture geometric nuances critical for obstacle-aware navigation. Combined with interleaved feedback, this enables the robot to incrementally build accurate geometric understanding, directly contributing to the consistent collision reduction across all scenes.

\textbf{Sensing Economy:}
Inter-POMDP consistently requires fewer detection activations. By the third object, it requires only a single detection across all scenes, compared to 2 (CSG-TL) and 1-4 (COSPOMDP). As shown in Figure~\ref{fig:simulation}, the aligned timestamps reveal that Inter-POMDP directs the robot to more precisely targeted locations, reducing exploratory sensing - an important advantage for real-world deployment where each detection incurs computational and time costs.

\textbf{Cross-Scene Generalizability:}
These trends hold consistently across three ProcTHOR scenes. Inter-POMDP leverages LLMs for semantic reasoning and maintain stable performance across scenes, whereas CSG-TL and COSPOMDP learned correlation model exhibit high variance (e.g., 108 vs. 359 navigation steps for object\_1 across train\_1 and train\_8). Additionally, model-based semantics alone are insufficient: CSG-TL's lack of geometric feedback still leads to suboptimal navigation. The combination of model-based semantic priors with interleaved geometric feedback, as in Inter-POMDP, generalizes most reliably across diverse environments without scene-specific tuning.


\subsection{Real-world Experiments}

\begin{figure*}
    \centering
    \begin{subfigure}[b]{\textwidth}
        \includegraphics[width=\textwidth]{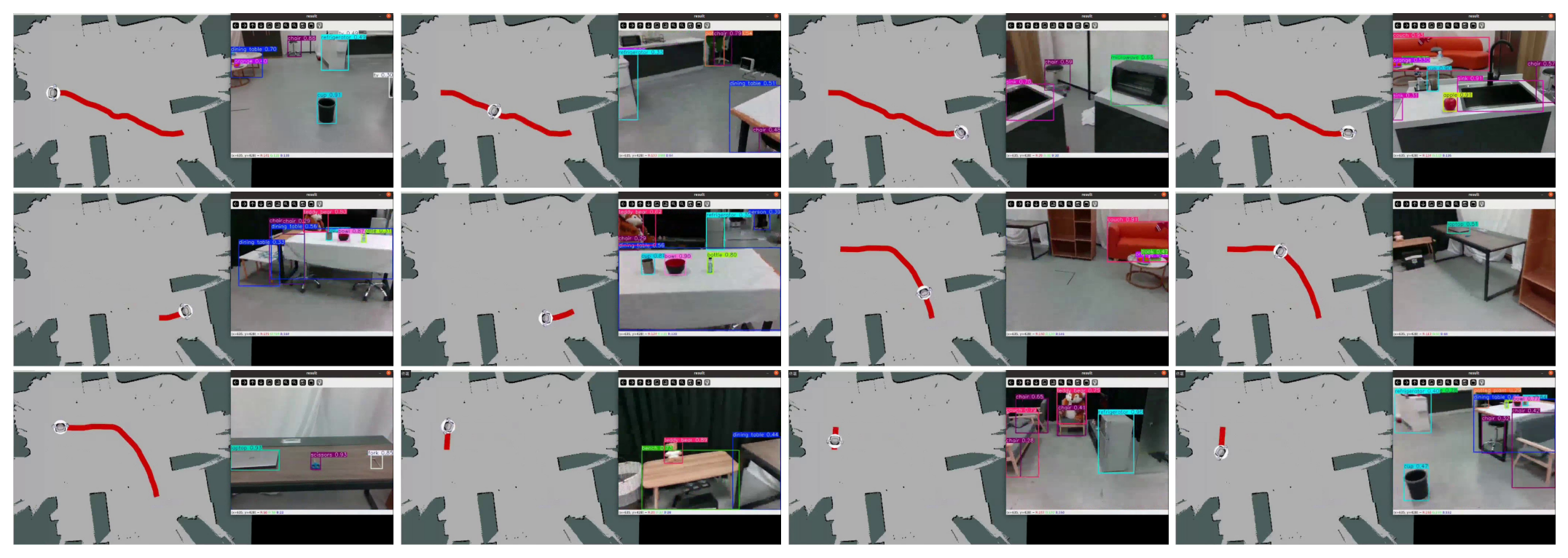}
        \caption{Top-down view of the planning interface. The red trajectory shows the robot's path, while the side window displays real-time semantic object detection results (e.g., apple, cup, fork) in the current camera frame.}
        \label{fig:experiments/real/breakfast_top}
    \end{subfigure}
    \hfill
    \begin{subfigure}[b]{\textwidth}
        \includegraphics[width=\textwidth]{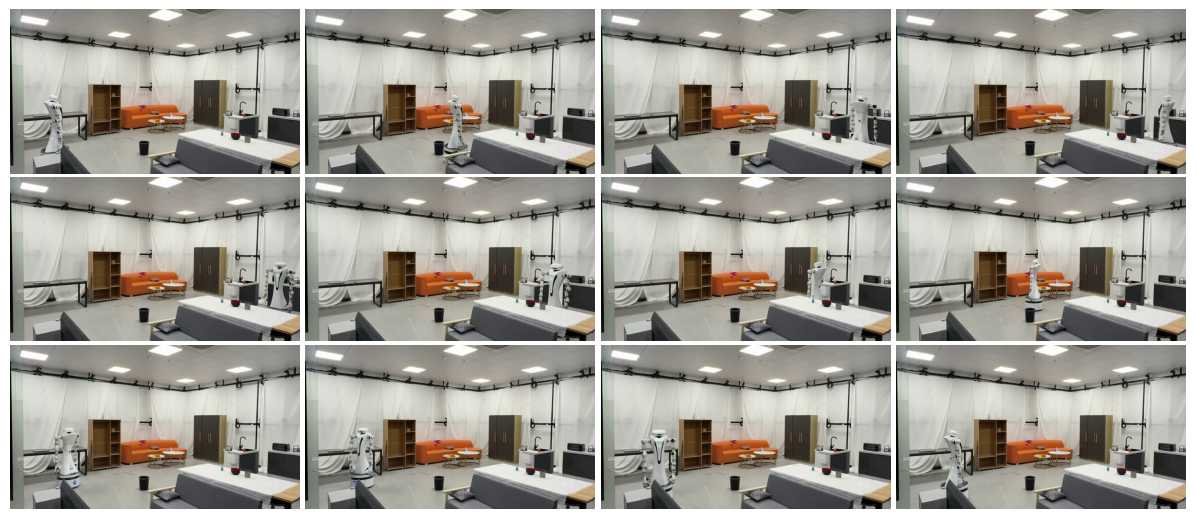}
        \caption{Third-person perspective of the robot traversing the multi-room environment to locate breakfast items. The sequence highlights the robot's navigation between topological nodes (e.g., dining table, sofa) as directed by Inter-POMDP.}
        \label{fig:experiments/real/breakfast_whole}
    \end{subfigure}
    \caption{Real-world experiments of searching apple/cup/fork in an unknown multi-room environment to prepare breakfast. Driven by our Inter-POMDP algorithm, the robot efficiently found three targets. Full videos are \url{https://sites.google.com/view/inter-pomdp}}
    \label{fig:real main}
\end{figure*}

We deploy Inter-POMDP on a real mobile manipulator tasked with finding a cup, apple, and fork to prepare breakfast in an unknown multi-room environment. As shown in Figure \ref{fig:real main}, the robot maintains a dual-layer mapping process: the high-level POUCT planner updates the semantic map $M^s_t$ using object-furniture relations to narrow the search space, while the low-level motion planner updates the topological map $M^p_t$ from LiDAR observations to refine navigation paths. Upon each camera observation $z_t$, both histogram and particle beliefs are updated simultaneously, enabling the low-level planner to propagate navigation costs back to the high-level planner for pruning redundant search branches.

\section{Conclusion}
\label{conclusion}

We present Inter-POMDP, a novel interleaved POMDP planning algorithm for multi-object search in unknown multi-room environments. By coupling a high-level semantic planner (histogram beliefs with LLM priors) with a low-level motion planner (particle-based beliefs), our framework captures both semantic and geometric uncertainties through principled interaction between the two levels. Both simulation and real-world experiments demonstrate that Inter-POMDP reduces collision counts by up to 63\%, navigation steps by up to 35\%, and detection counts by up to 32\% compared with baselines. Future works could be: 1) extending to 3D environments for more realistic household scenarios \cite{zheng2021multi, zheng2023asystem}; 2) incorporating object manipulation in cluttered tabletop workspaces, where geometric uncertainties increase by orders of magnitude; 3) modeling continuous actions via POMCPOW \cite{sunberg2018online} for finer navigation, with learned samplers such as conditional VAEs \cite{sohn2015learning} to maintain planning efficiency.


\newpage

\bibliography{IEEEabrv.bib}
\bibliographystyle{IEEEtran}

\end{document}